\documentclass[final,3p,times]{elsarticle}

\usepackage{framed,multirow}
\usepackage{lineno}
\usepackage{amssymb}
\usepackage{amsmath}
\usepackage{latexsym}
\usepackage{setspace}

\usepackage{url}
\usepackage[dvipsnames]{xcolor}
\definecolor{newcolor}{rgb}{.8,.349,.1}
\usepackage[utf8]{inputenx}
\usepackage{graphicx}
\usepackage{amsmath,amssymb,mathrsfs}
\usepackage{fancyhdr}
\usepackage{calc}
\usepackage[T1]{fontenc}
\usepackage{mathptmx}
\usepackage{physics}
\usepackage{lineno}
\usepackage{tikz}
\usepackage{pgfplots}

\newcommand{\bes}{\begin{equation*}}
\newcommand{\ees}{\end{equation*}}
\newcommand{\beq}{\begin{equation}}
\newcommand{\eeq}{\end{equation}}
\newcommand{\bea}{\begin{eqnarray}}
\newcommand{\eea}{\end{eqnarray}}
\newcommand{\beas}{\begin{eqnarray*}}
\newcommand{\eeas}{\end{eqnarray*}}

\begin{document}

\begin{frontmatter}

\title{ONERA’s CRM WBPN database for machine learning activities, related regression challenge and first results}

\author[rvt1]{Jacques Peter\corref{cor1}}
\ead{jacques.peter@onera.fr}
\cortext[cor1]{Corresponding author. Tel.: +33 1 46 73 41 84.}
\author[rvt2]{Quentin Bennehard}
\author[rvt1]{Sébastien Heib}
\author[rvt2]{Jean-Luc Hantrais Gervois}
\author[rvt2]{Frédéric Moëns}
\address[rvt1]{DAAA, ONERA, Universit\'e Paris Saclay, F-92322 Ch\^atillon, France}
\address[rvt2]{DAAA, ONERA, Institut Polytechnique de Paris, 92190, Meudon, France}

\begin{abstract}

This paper presents a new Computational Fluid Dynamics database, developed at ONERA, to support
the advancement of machine learning
   techniques for aerodynamic field prediction. It contains 468 Reynolds-Averaged Navier-Stokes simulations using the Spalart-Allmaras
  turbulence model, performed on the NASA/Boeing Common Research Model wing-body-pylon-nacelle configuration. The database spans a
  wide range of flow conditions, varying Mach number (including transonic regimes), angle of attack (capturing flow separation), and Reynolds number
  (based on three stagnation pressures, with one setting matching wind tunnel experiments).
  The quality of the database is assessed, through checking the convergence level of each computation.

  Based on these data, a regression challenge is defined. It consists in  predicting the wall distributions of pressure and friction coefficients
   for unseen aerodynamic conditions.
   The 468 simulations are split into training and testing sets, with the training data made available
   publicly on the Codabench platform.
  The paper further evaluates several classical machine learning regressors on this task.
  Tested pointwise methods include Multi-Layer Perceptrons, $\lambda$-DNNs,
  and Decision Trees, while global methods include Multi-Layer Perceptron, k-Nearest Neighbors,
  Proper Orthogonal Decomposition and IsoMap.
  Initial performance results, using $R^2$ scores and worst relative mean absolute error metrics,
  are presented, offering insights into the capabilities of
  these techniques for the challenge and references for future work.

\end{abstract}


\end{frontmatter}


\section{Introduction}
The first purpose of this paper is to describe a database of Computational Fluid Dynamics (CFD)
simulations, recently built at ONERA, to support the development and validation of machine learning
techniques in the field of surface
field generation. A consistent set of simulations  about the NASA/Boeing Common Reserch Model (CRM)
aircraft has been run recently by
the authors. The retained geometry includes the wing, the body, the pylon and the nacelle \cite{TinBroKey_18}.
The selected model is the Reynolds-Averaged Navier-Stockes (RANS) set of equations completed by the Spalart-Allmaras (SA) turbulence models
$n_f=468$ flow conditions have been considered. The database  encompasses Mach number effects
(including the transonic regime),
incidence effects
(including flow separations) and Reynolds number (Re) effects (with transition fixing).
Its features and verification are detailed in section 2.
\\
\indent Based on these wall distributions,  an open access regression challenge is proposed.
Two thirds of the wall distributions are retained
for training and validation/selection of hyperparameters whereas one third  is used to test the regressors accuracy.
The provided inputs for  pointwise regressors are  the coordinates of the wallpoints, the local normal and the flow conditions.
Classical modewise regressor use only the flow conditions as input (and predict the complete
distribution of output variables) that can be extracted
from the latter set of inputs. The quantity of interest are the pressure coefficients and the
three components of the friction constraint.
A detailed presentation of the data, their splitting and the regression metrics is given in section 3.
\\
\indent Currently, the cost of a series of aerodynamic simulations with a high fidelity model and a dense mesh remains high
 just as, typically, an aircraft programm requires accurate data all over the flight envelope.
Therefore, there is a continuous interest in regressors predicting unseen wall distributions from
known distributions for a set of flow condtions or design/control parameters. Accurate wall distribution predictions
are particularly valuable in aeroelastic and aero-structure computations and optimizations \cite{SomMicLin_16}. Furthermore,
integrating these distributions yields aerodynamic coefficients, enabling the evaluation of aircraft performance
across diverse mission scenarios \cite{LieMadMar_15}.
\\
\indent Currently, a variety a methods are used. Some of them define a pointwise regressor.
Numerous alternative approaches predict complete
distributions of the quantities of interest
and will be designated as global regressors.  To get insight in the complex
landscape of candidates for wall distribution regression,
the interested reader is referred to the relevant sections of recent review articles
\cite{YonAndVal_18,LiDuMar_22,LeCFerGib_23}
and to these research reports or articles \cite{SabStuBek_22,HinBek_23,And_24}.
In section 4 and 5, classical Machine Learning (ML)
 techniques are briefly described as well as their accuracy for the regression challenge of interest.  
%
%
%
\section{Wing-body-pylone-nacelle CRM data base }
\subsection{Geometry}
The configuration supporting this activity is the Common Research Model (CRM) shared by NASA for
the AIAA drag prediction workshops. The selected version of the model features a fuselage, a wing and a
through flow nacelle-pylon. It has been used in the 6th edition of the drag prediction workshop
\cite{TinBroKey_18,HueChaLan_18}. This
configuration benefits from many simulations and wind tunnel measurements. NASA conducted several tests on
it and others conducted tests on more or less complete CRM geometries (without the nacelle but with vertical
fin in the ONERA S1MA wind tunnel, for instance \cite{CarHueCha_18}).
Whereas the configuration is representative of a transport aircraft cruising at Mach number 0.85 for a
Reynolds number of 40 $\times 10^6$, the shared data from the experiments are acquired for a Reynolds number
near 5 $\times 10^6$, that has been retained as one the Re of the computational basis.
%
%
\subsection{Selected flow conditions}
\begin {figure}[htbp]
  \begin{center}
	  \includegraphics[width=0.6\linewidth]{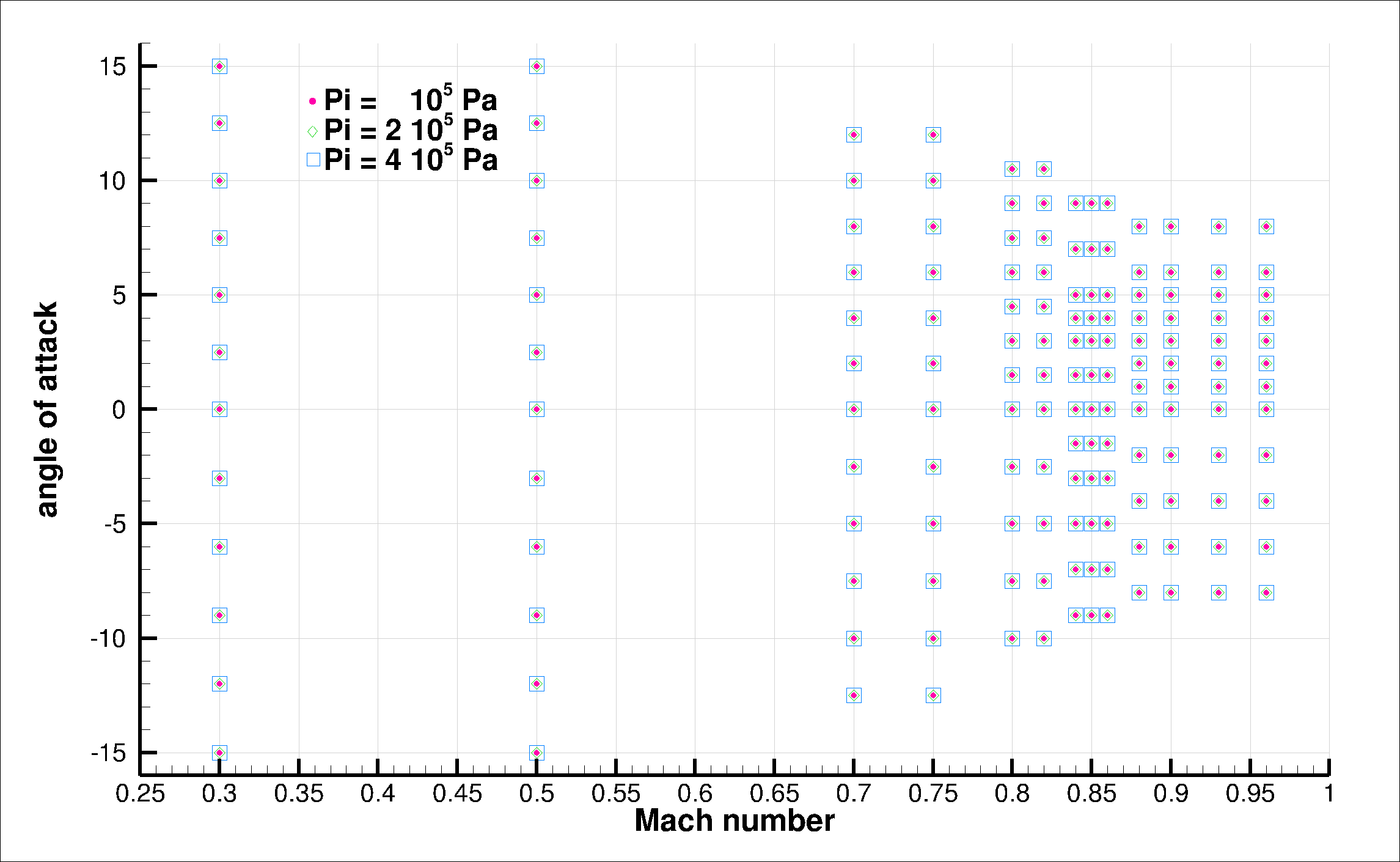}
	  \caption{Numerical Design Of Experiment}
  \label{fig_numdoe}
  \end{center}
 \end{figure}

The numerical Design of Experiments (DoE), presented in Figure \ref{fig_numdoe}, includes 13 far-field Mach numbers ($M_\infty$ )
values from 0.30 do 0.96 and, for each of them, 12 angles of attack (AoA) in decreasing intervals as $M_\infty$ 
increases -- precisely from AoA in [-15$^o$ ,15$^o$ ] for $M_\infty$ = 0.3 and 0.5 to
AoA in [-8$^o$, 8$^o$ ] for $M_\infty$ = 0.88 to 0.96. To allow for Reynolds number
(Re) effects, typically between various wind tunnels, we have considered three stagnation states:  a unique
realistic stagnation temperature with three stagnation pressures $p_i = 10^5$, $ 2 \times 10^5$ or $4 \times 10^5$ Pa.
 A set of variables  ($M_\infty$, AoA, $p_i$) fully defines a flow and is denoted in short as $\bf{p}$.    
\\
\indent The usual Reynolds number for external flows is defined from the velocity and the thermodynamic variables
about the solid shape. Using the classical laws for perfect gaz
and  isentropic-isenthalpic trajectories
\beq
Re = \frac{\rho ~ V  ~L }{\mu(T) } =   \frac{ p ~ M~ c ~ L }{r~ T ~\mu(T) } = \sqrt{\frac{\gamma}{r}} \frac{ p ~ M~   L }{\sqrt{T} \mu(T) } =
\sqrt{\frac{\gamma}{r}} \frac{ p_i ~ M~   L }{\sqrt{T_i} ~\mu(T)} (1+\frac{\gamma-1}{2}M^2)^{\frac{1-2 \gamma}{2(\gamma-1)}},  \label{eq_Re1}
\eeq
    denoting $(\rho,p,T)$ the density, the static pressure and the temperature, $(\rho_i,p_i,T_i)$ the corresponding stagnation
    quantities, $V$ the velocity, $\mu$ the molecular viscosity, $\gamma$ the ratio of specific heats,
    $r$ the constant of the law of perfect gaz
    and $L$ the CRM reference length. Finally, the viscosity as a function of the temperature
    is defined by Sutherland's law,
\beq
\mu(T) =\mu(T_{ref})\left(\frac{T}{T_{ref}}\right)^{3/2}  \frac{T_{ref}+S}{T+S}    \qquad \left(  \textrm{with} \quad
  T = T_i (1+\frac{\gamma-1}{2}M^2)^{-1}) \right)
\label{eq_Re2}
\eeq
with a set of suitable constants $(T_{ref},\mu(T_{ref}),S)$ for air. 
 From equations (\ref{eq_Re1}) and (\ref{eq_Re2}), it is clear that, for each ($M_\infty$, AoA) couple, with a unique stagnation temperature
    $T_i$ and stagnation pressures $p_i$ in the proportion of 1, 2 and 4, 
 the  Re of the flows at the same Mach number in the three wind tunnels are also in the proportion of 1, 2 and 4.
For the $M_\infty$ = 0.85 flows, these Reynolds numbers are precisely Re = 2.5 $\times 10^6$, 5
$\times 10^6$ and 10 $\times 10^6$. The second flow studied during the 6th Drag Prediction Workshop (DPW6),
at $M_\infty$ = 0.85 and Re = 5 $\times 10^6$ about the CRM,
is included in the database or, more precisely, a target lift flow
was considered during DPW6 about the same geometry
and at one of the farfield Mach and Reynolds numbers
 of the database \cite{TinBroKey_18}.
 \\
  \begin {figure}[htbp]
  \begin{center}
	  \includegraphics[width=0.48\linewidth]{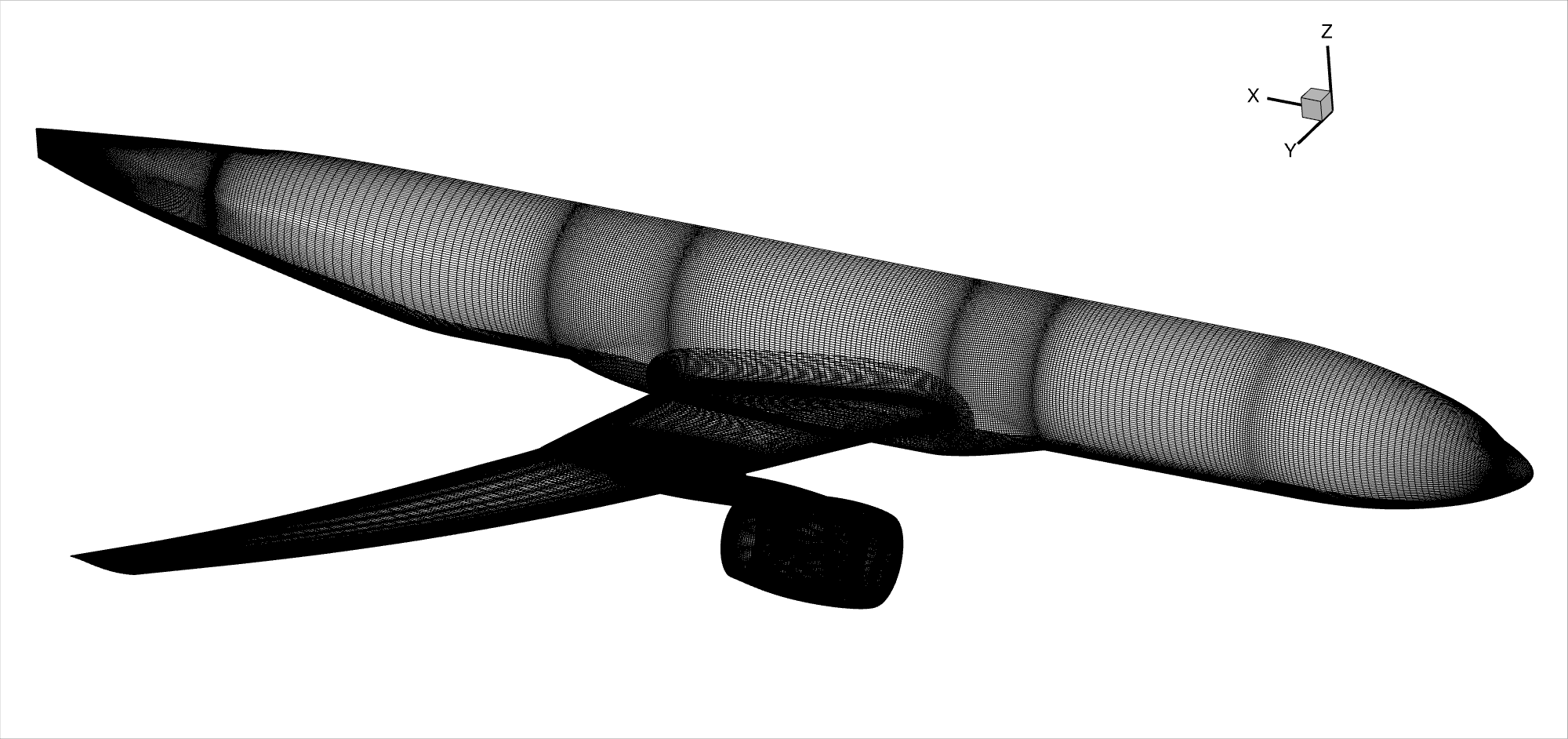}
	  \includegraphics[width=0.48\linewidth]{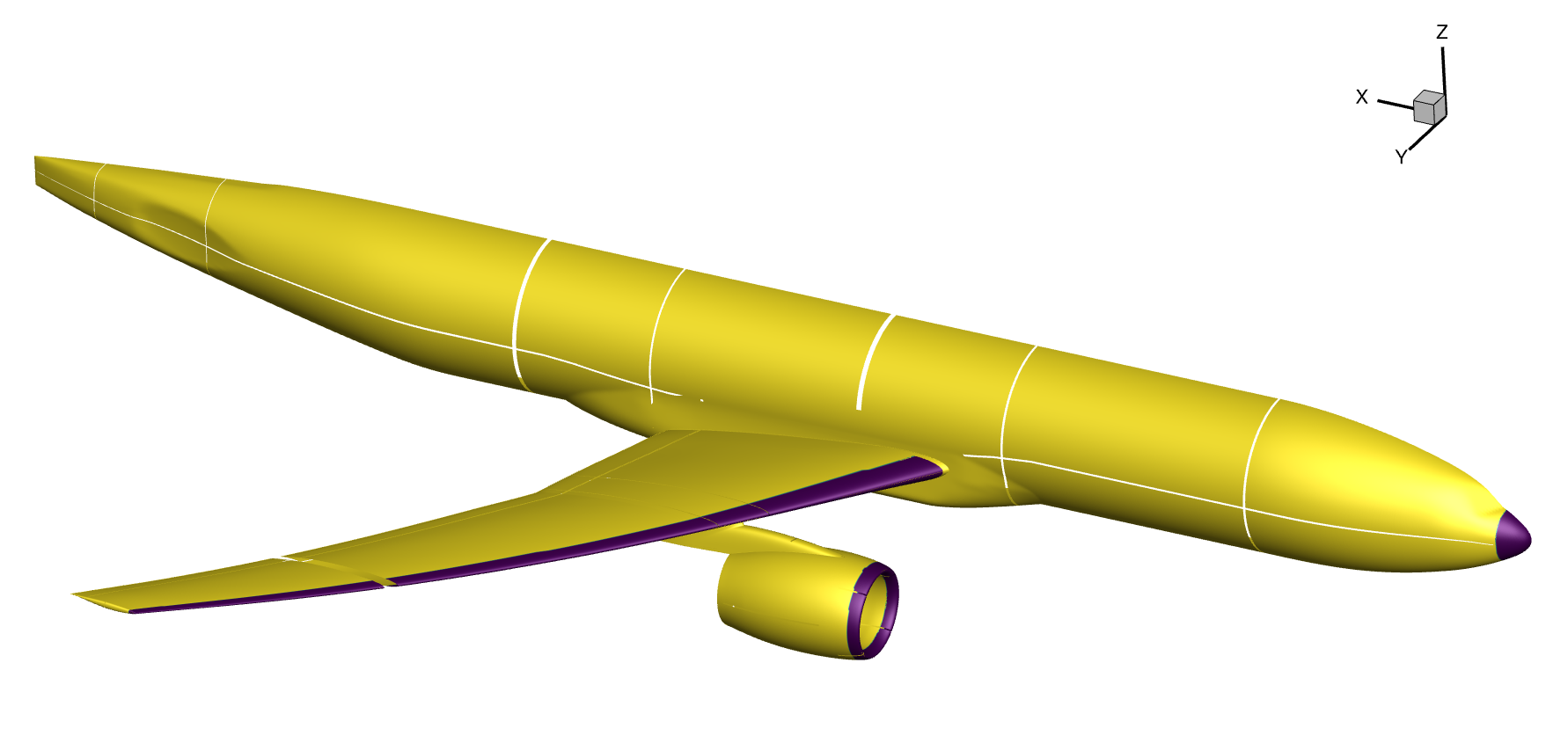}
	  \caption{CRM WBPN. Left: wall mesh. Right: imposed laminar zones}
  \label{fig_wallmesh_interm}
  \end{center}
\end{figure}
 %
%
\subsection{CFD simulations}
The 468 RANS numerical simulations have been run with the ONERA elsA software \cite{CamHeiPlo_13}.
A Boeing structured Chimera mesh including 39.5M
cells has been selected from the DPW database  -- see (WBPN3) in [2] and Figure \ref{fig_wallmesh_interm}
left -- leading to satisfactory
$y^+$ values for all computations. The elsA software solves the compressible 3D RANS equations by using a cell-centered
finite volume spatial discretization on the structured multi-block mesh.
\\
\indent Computations are carried
out using an uncoupled approach between the RANS system and the turbulence model transport equations.
Regarding the mean flow, a space centered type scheme, as proposed by Jameson et al., is retained for the
conservative variables \cite{JamSchTur_81}. A fourth order linear dissipation is used (coefficient $k_4$ ), with added second
order dissipation terms (coefficient $k_2$ ) for the treatment of flow discontinuities. Although calculations at high
positive and negative AoA are difficult to converge and may even not admit a steady-state solution, a unique set
of low values ($k_2$, $k_4$)=(0.50,0.16) has been used for the numerical dissipation parameters. The viscous fluxes
use a classical formulation with a 3-point stencil per mesh direction based on face-centred Green-Gauss gradients.
\\
\indent The selected turbulence closure is the Spalart-Allmaras model with Quadratic Constitutive Relation (QCR) \cite{Spa_00}.
For the convection term, Roe's numerical scheme with Harten's entropy correction is used \cite{Roe_81,Har_83}. The diffusive term
involves, as for the mean flow, face-centred gradients whereas the source terms are based on cell-centred
gradients.
\\
\indent A LUSSOR implicit stage is associated with an Euler backward time integration scheme, which allows fast
convergence rates \cite{YooJam_88}. Finally, a local time stepping and multigrid techniques based on a V-cycle with one coarse
grid are applied to speed up the convergence process [7].
\\
\indent Laminar zones at the wall and its vicinity have been imposed consistently upwind the transition strips used
in the ONERA's experiments \cite{CarHueCha_18}. More precisely, the strips were installed at 10\% chord on the wings and at
1.6\% of the total fuselage length downstream the nose.
The wall areas upstream the strips appear in violet in the right part of Figure \ref{fig_wallmesh_interm}. The turbulent viscosity is set to zero during the
computations for all points of the fluid domain which closest wall point belongs to these zones coloured in
violet. The engines contribution is not simulated (the nacelles are considered through flow).
%
%
%
\subsection{Numerical improvement and verification of the database}
\begin {figure}[htbp]
  \begin{center}
	  \includegraphics[width=0.58\linewidth]{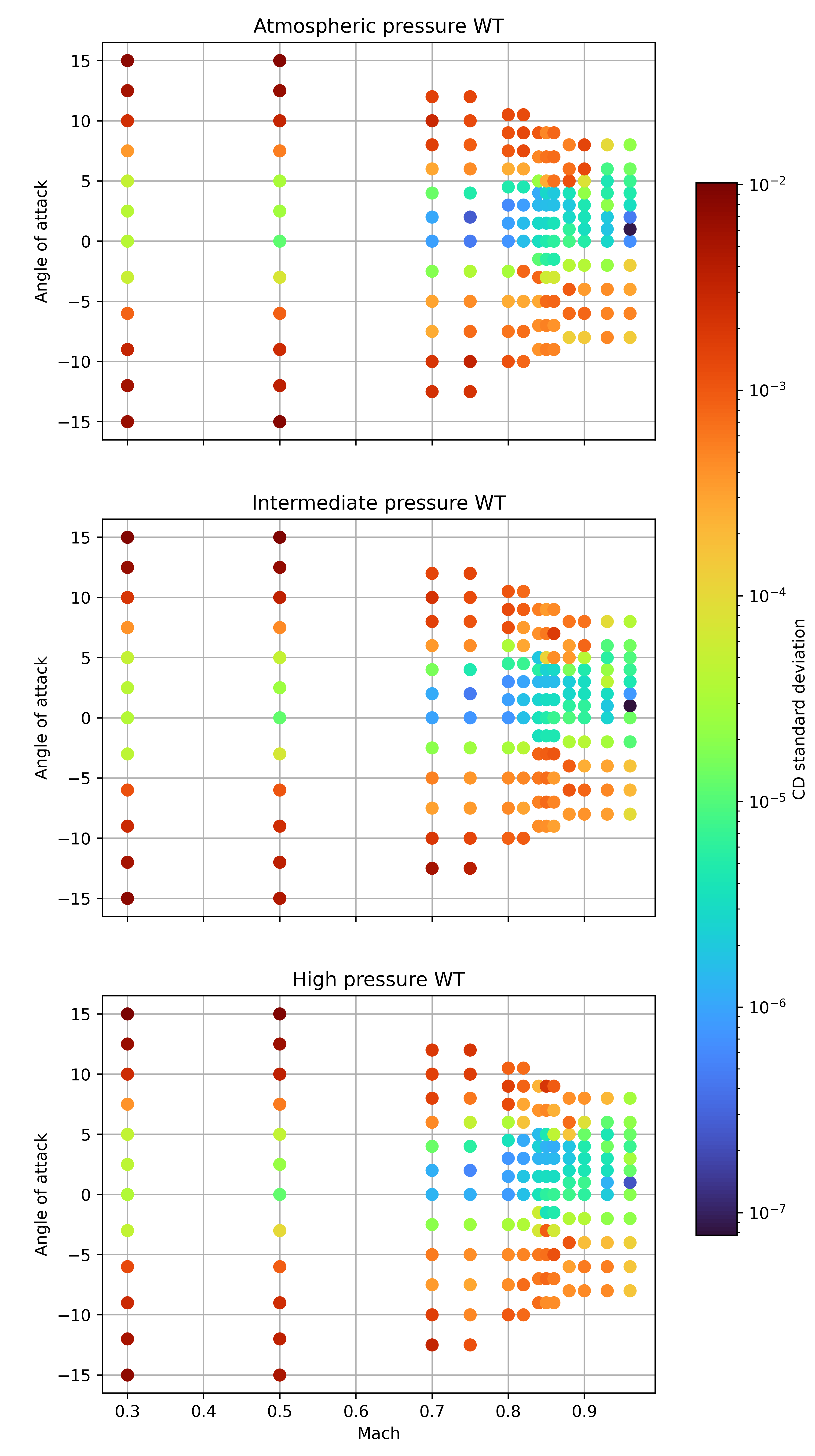}
	  \includegraphics[width=0.41\linewidth]{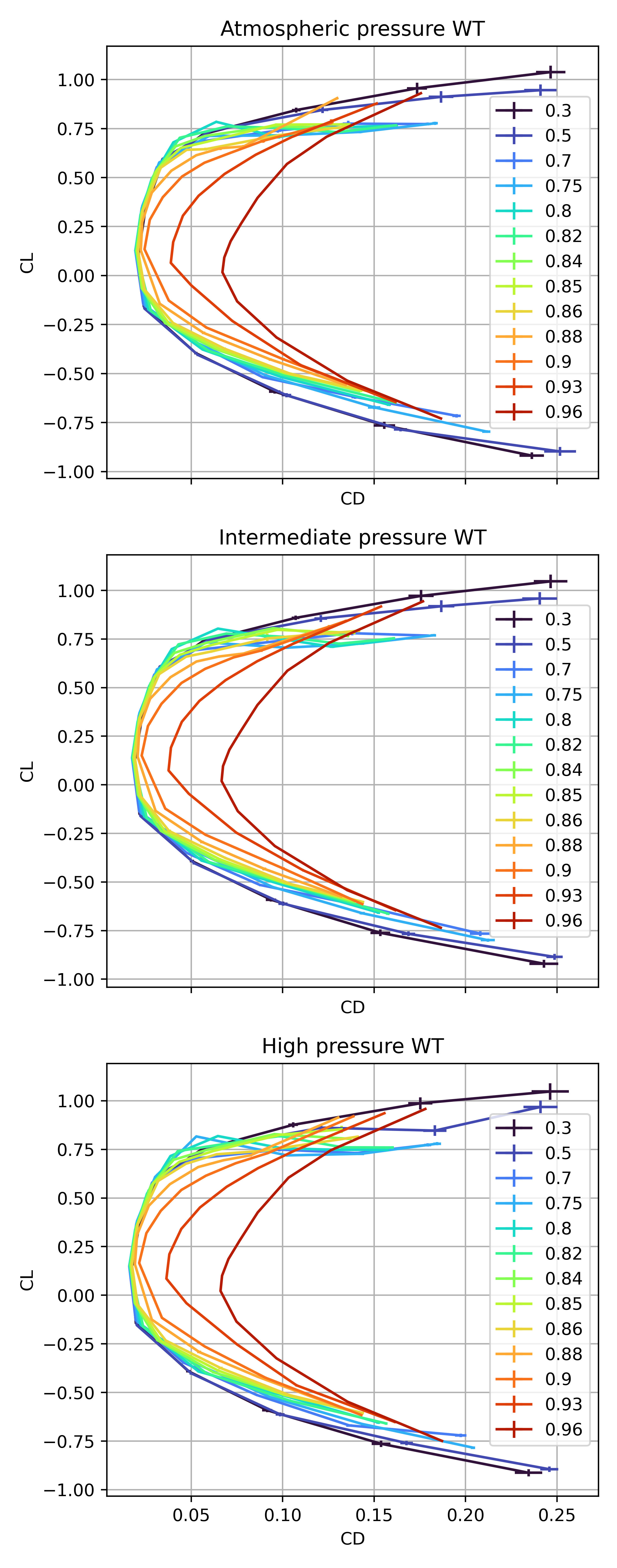}
	  \caption{Left: Numerical verification, standard deviation of the drag over the last 20\% iterations.
      Right: For drag and lift, relative importance of change in flow
      conditions and standard deviation }
  \label{fig_cdstd}
  \end{center}
\end{figure}

For each computation, the standard deviation of the drag over the last 20\% iterations as well as the explicit
residual history have been extracted. These two metrics can be used as confidence factors, informing the database user
about the quality of the calculations. After the complete series of computations, the drag standard deviation
of a few points of the DoE appeared to be high compared to the one of their neighbors. A specific effort was devoted
to these points testing specific numerical options, like the increase of coarse grid numerical dissipation or an additional
fine grid iteration after the standard V-cycle, without ever altering the numerical scheme on the fine grid numerical scheme.
This task has significantly improved the convergence metrics of three computations in the high-$p_i$ part of the DoE.    
\\
\indent The drag standard deviation after the control
and improvement step is given in classical non-dimensional value for all simulations  --
see Figure \ref{fig_cdstd}.
The convergence of the calculations is judged very satisfactory at
moderate angles of attack – AoA in [-3$^o$, 3$^o$] – for all the Mach and Reynolds numbers of the database.
Low values below 10$^{-4}$ (one drag count) can be achieved for these flow conditions.
For lower or higher angles of attack, the stability and convergence
metrics are degraded due to the non-linear aerodynamic flows, as can be expected from RANS simulations. It is
especially true at low Mach number ($M_\infty$ = 0.3 and $M_\infty$ =
0.5) and extreme values of angle of attack (above 10$^o$ and below -10$^o$).
\\
 \begin {figure}[htbp]
  \begin{center}
	  \includegraphics[width=0.6\linewidth]{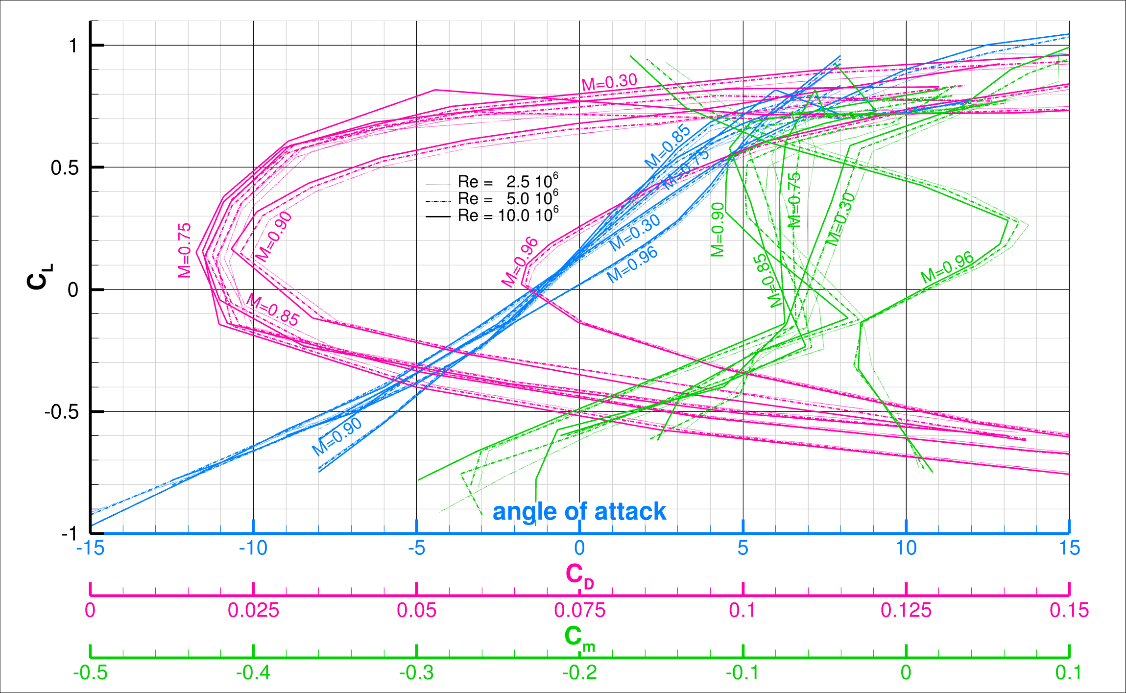}
	  \caption{Example of polar plots from the database (at various Mach and Reynolds numbers)}
  \label{fig_clcdcm}
  \end{center}
 \end{figure}
 \begin {figure}[htbp]
  \begin{center}
	  \includegraphics[width=0.32\linewidth]{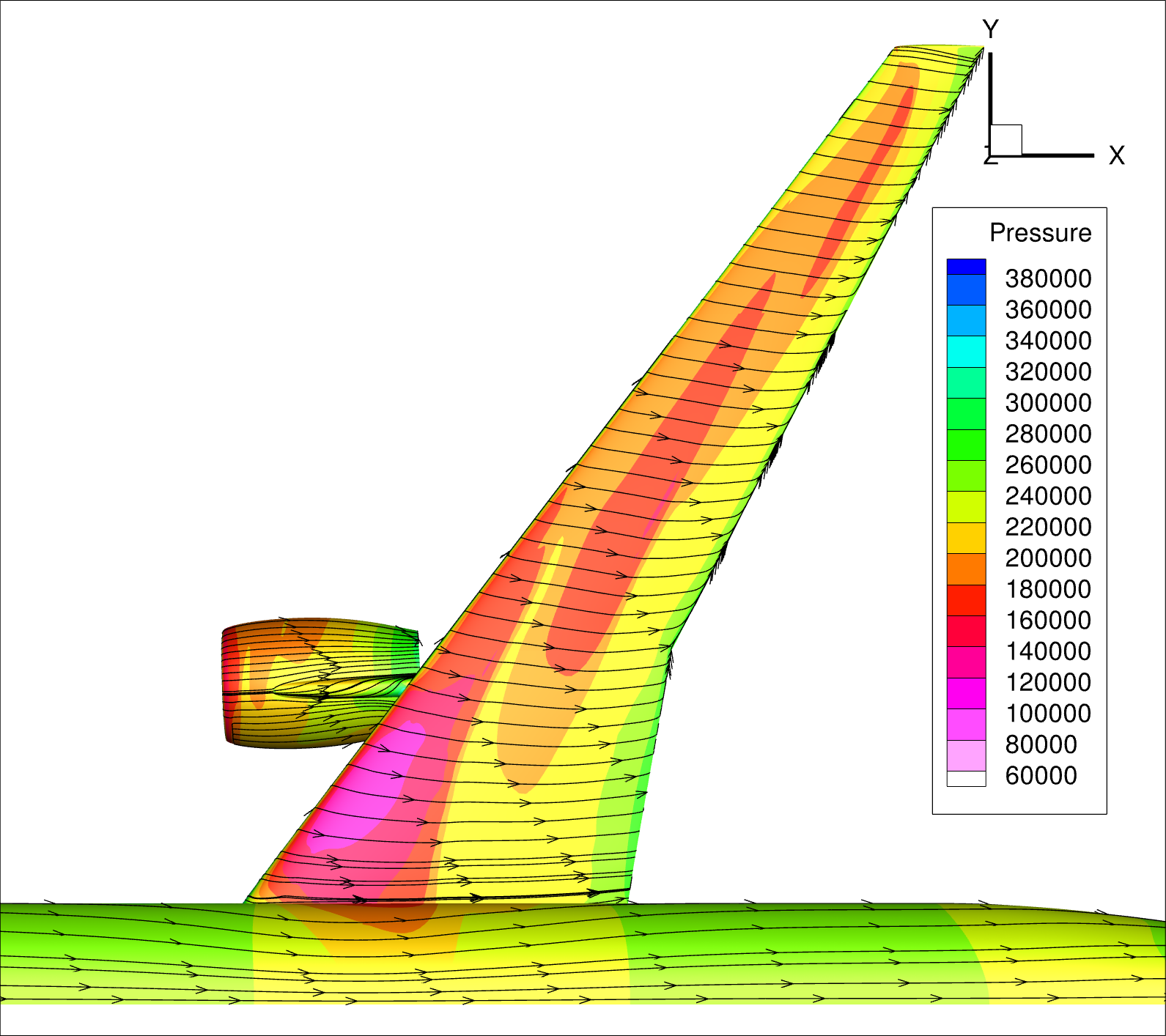}
	  \includegraphics[width=0.32\linewidth]{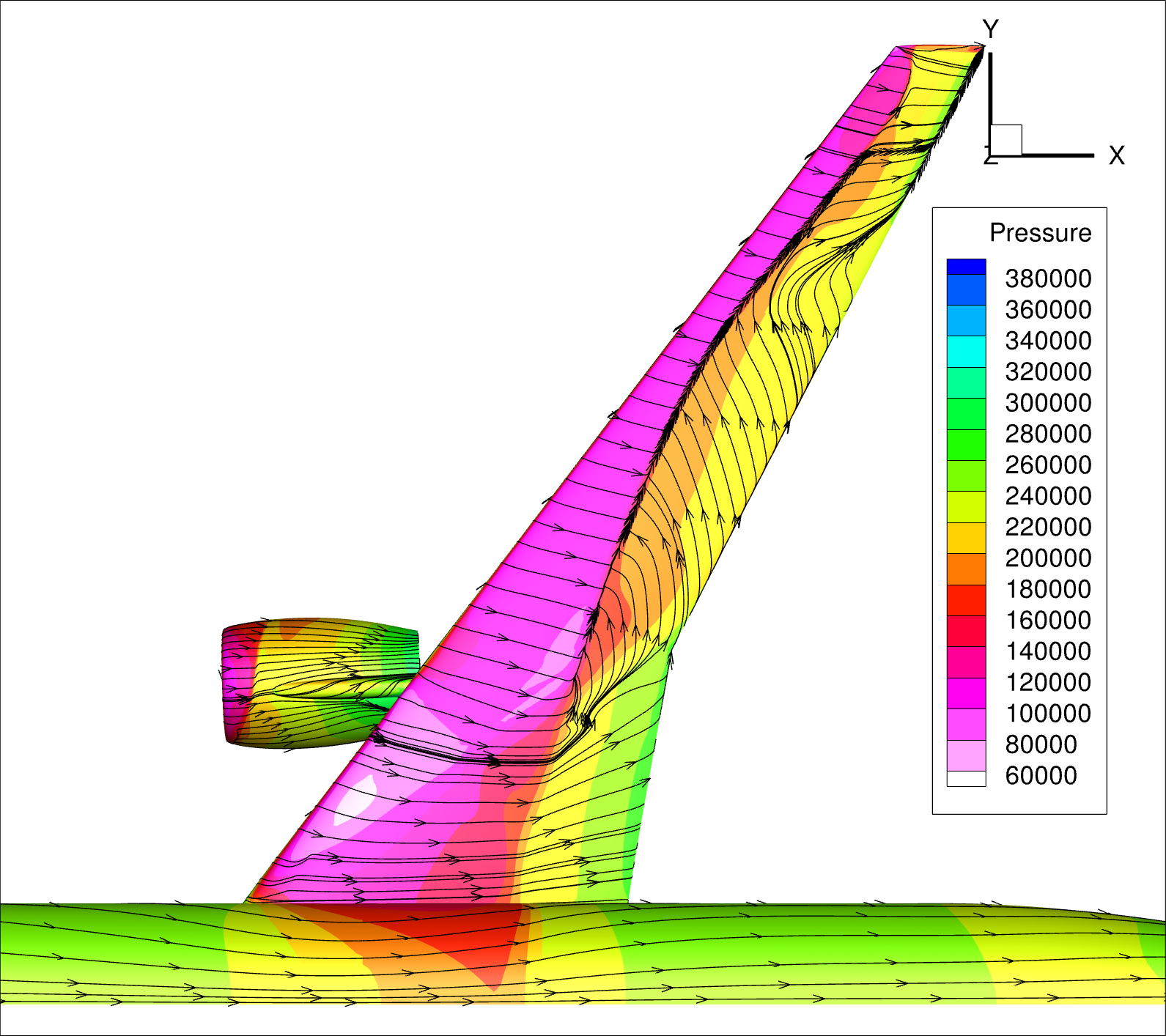}
	  \includegraphics[width=0.32\linewidth]{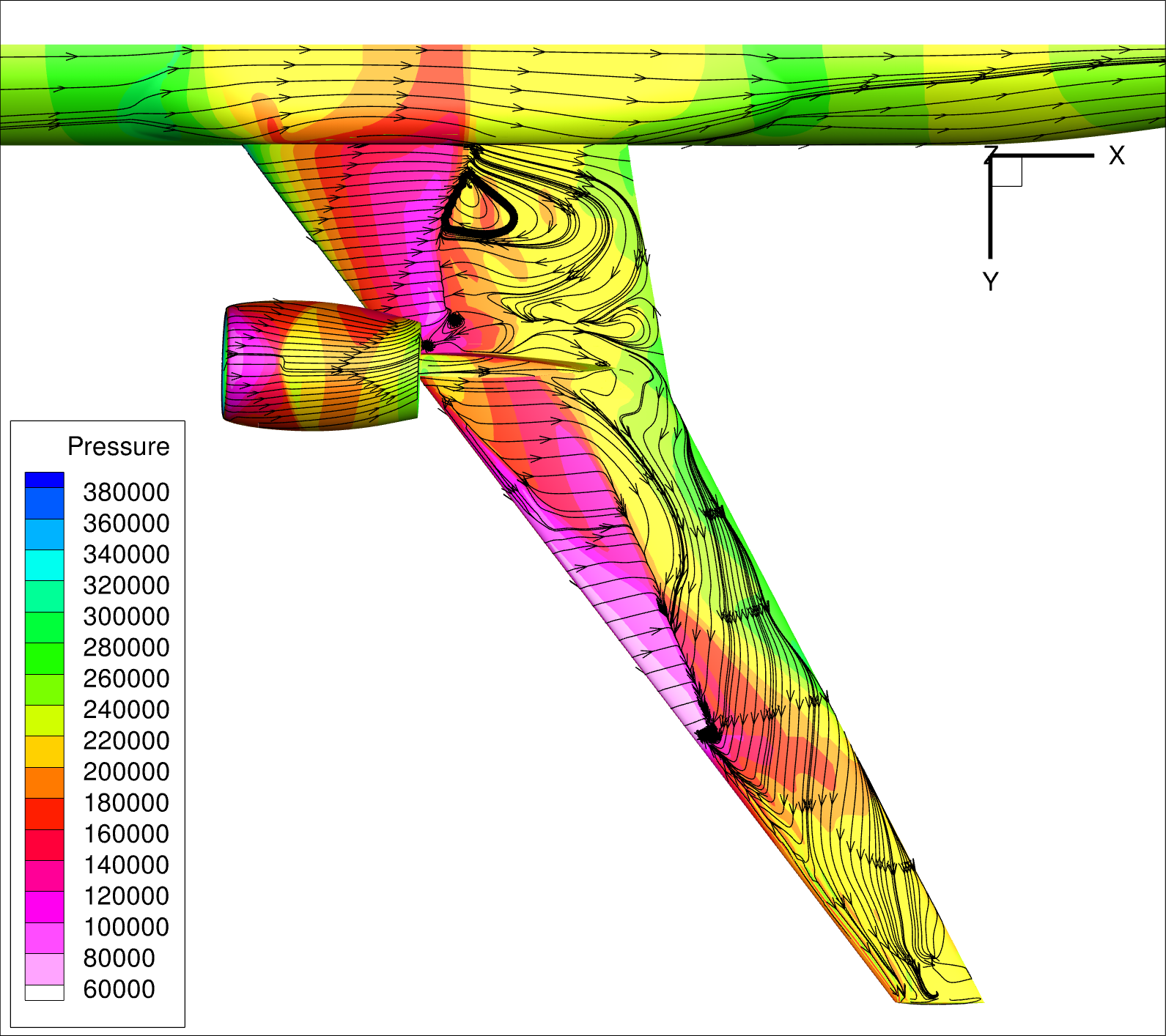}
	  \caption{isolines of pressure and friction lines for $p_i = 4 \times 10^5$ (all) and $M_\infty = 0.85$,  AoA = 1.5$^o $  (left),
    AoA = 5$^o$ (center), AoA = -5$^o$ (right).}
  \label{fig_presfricMinf085}
  \end{center}
 \end{figure}
 %
%
\subsection{Physical verification of the database}
The outcome of the simulations consists of the forces and moments coefficients -- see Figure \ref{fig_clcdcm}
-- and the stresses at the model
surface (pressure and friction components). At a global level, the convergence of the total force coefficients
across the database is quite satisfying (incidence and Mach number effects). Also the expected Reynolds number effects
can be observed on the drag: when the Reynolds number increases, the total drag decreases due to the viscous drag
reduction. 
\\
\indent A wide range of aerodynamic
relevant features are covered. Low speed pressure results are represented together with transonic flow
fields. Several examples of the pressure and friction fields are shown in Figure \ref{fig_presfricMinf085}.
 The figure exhibits a cruise condition with a mild shock and attached flow. When the incidence is
increased, stronger shocks occur on the wing upper surface and cause the flow to separate down to the
trailing edge. More complex shock and boundary layer flow separations are obtained on the lower wing for
negative angles of attack. The validity of these simulations for separated flows is inherently limited due
to the RANS approach and the turbulence model. 
Nonetheless, one can check that the spacing between two conditions (like two adjacent angles of attack)
results in a significantly greater gap between the CL and CD values, that what is caused by the relatively bad
convergence of some extreme points – see Figure \ref{fig_cdstd} right.
%
%
\subsection{Scattered  and structured data}
\begin {figure}[htbp]
  \begin{center}
	  \includegraphics[width=0.48\linewidth]{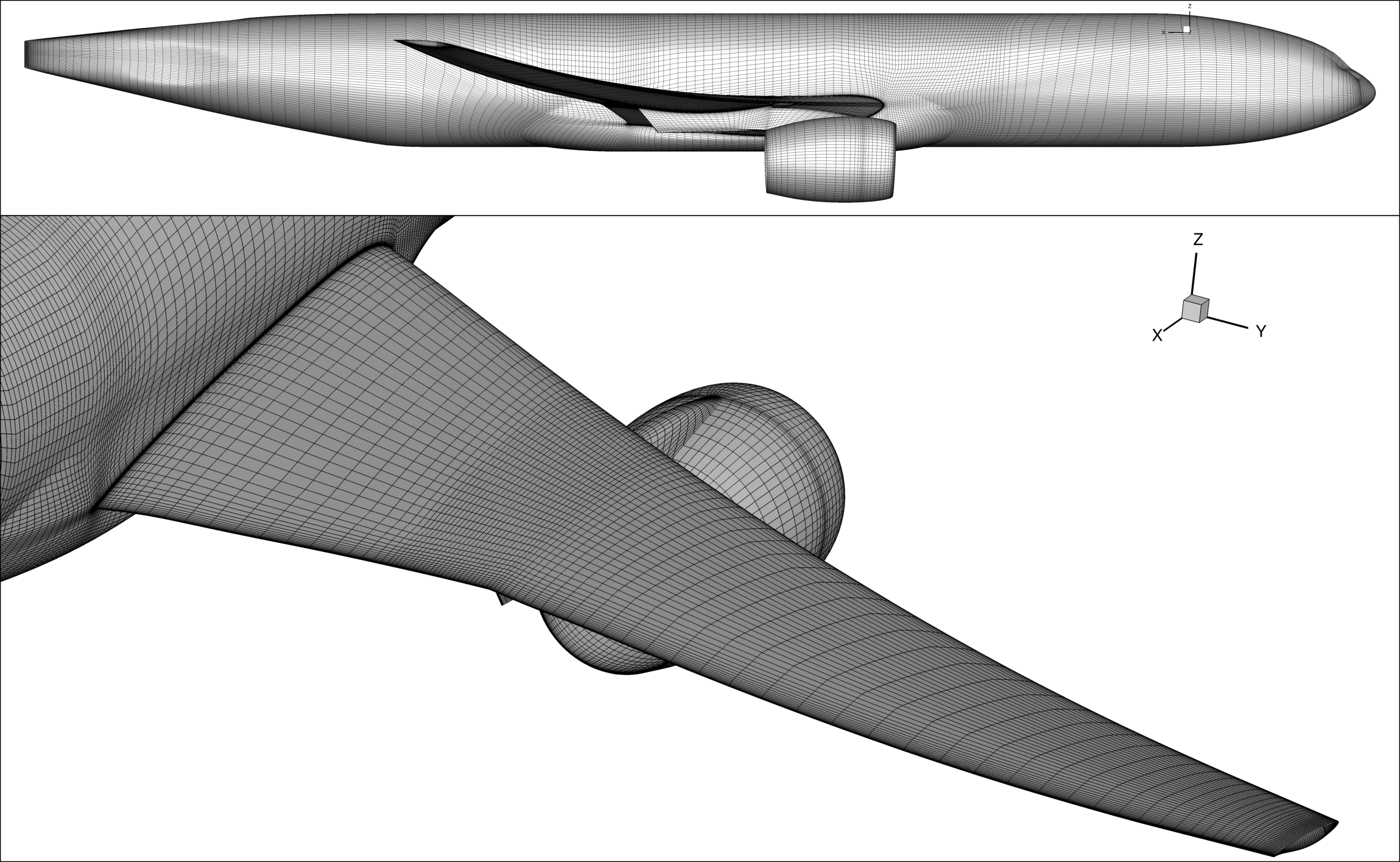}
	  \includegraphics[width=0.48\linewidth]{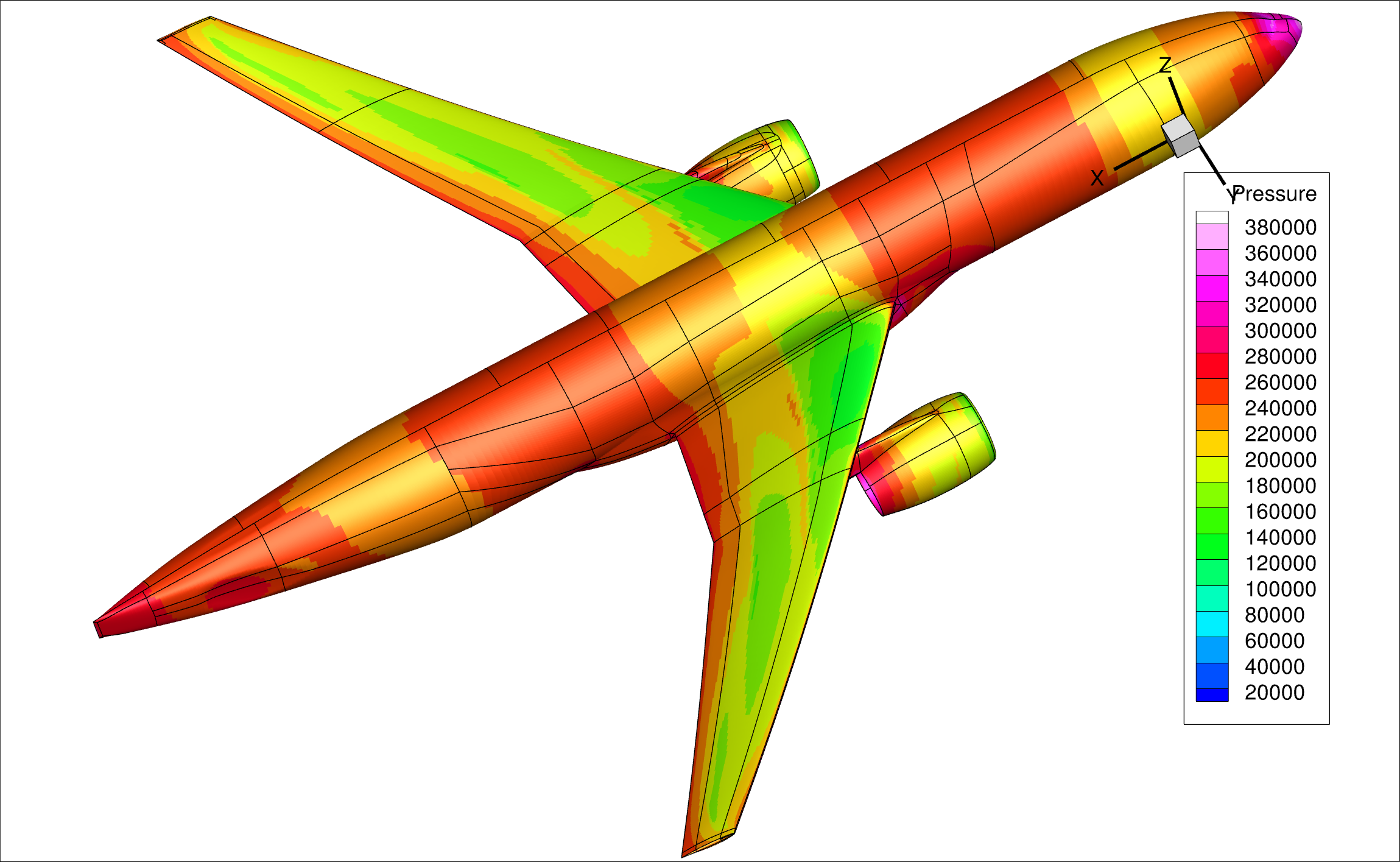}
	  \caption{CRM WBPN. Left: Structured surface mesh shared on the drag prediction workshop database  (to allow for visualization, every other node is skipped in each direction).
      Right: Chimera solver output (port wing) compared to the interpolation on the structured mesh (starboard wing)}
  \label{fig_wallproj}
  \end{center}
\end{figure}

The stored outputs of the data basis are the flow variables at the wall
and the classical forces and moments. The wall distributions, as outputs of the computations, are
available on overlapping patches as some of the
Chimera zones are adjacent to the wall. They have been converted in pointwise data using a classical priority
criterion in overlapping zones. The size of the resulting pointwise fields is $n_p=260,774$. Comparison of isolines
between structured-CFD plots and scatter plots provided a complete validation for the pointwise fields.
\\
\indent The surface data have also been projected on a non- overlapping multi-patch structured wall-mesh (see
Figure  \ref{fig_wallproj} left), given in the frame of the DPW6 by Charbonnier \cite{TinBroKey_18}.
The mesh being coincident
and structured, topological machine learning might be considered in the future using these specific transformed
data. This structured mesh used for interpolation features $n_{pi}=188,895$ nodes and can be simplified
by taking only half the nodes in each
direction (47,351 nodes as in the picture). High order interpolation is used for the projection on the structured
mesh (see Figure \ref{fig_wallproj} right). In order to validate the precision of these interpolated fields, the resulting
pressure and friction fields have been integrated and compared to the total coefficients obtained by the
solver. The surface integration of the interpolated fields differs from the solver output by 1 $\times 10^{-4}$  to 5 $\times 10^{-4}$  in
drag. The surface integration outside the solver itself introduces some error (well below 10$^{-4}$ in drag, except
for 5 extreme conditions). The complete flow fields of the fluid domain were
stored on hard disk and may be used for future activities.
%
 %

%
\section{Regression challenge}
An aircraft program requires aerodynamic data all over the flight envelope and CFD is more and more involved
in the production of these data.
However, on top of possible accuracy issues, high fidelity aerodynamic and aeroelastic simulations about an
aircraft are still expensive and a complete design
of experiment for the flight envelope would be prohibitive. There is hence a strong interest in regression
methods predicting flows for unseen parameters
from a database of known numerical flows. The relevant sections of classical review articles may be used
as introductions to this field \cite{YonAndVal_18,LiDuMar_22}.
Besides, well-illustrated contributions appear in a recent GARTEUR group report \cite{And_24}  and in many conference papers
–- see for example \cite{HinBek_22,LipJP_24}.
\\
 \indent The pressure coefficient ($Cp$) and the
friction coefficient along each axis ($Cf_x$, $Cf_y$, $Cf_z$)  at the skin of the aircraft have been selected
as the outpout quantities of interest of the regression exercise.
 The scattered fields (size $n_p$), directly extracted from the CFD computations,
 have been retained. At each of the skin points, the coordinates in the cartesian frame of reference
 ($x$, $y$, $z$) and the components of
the normal vectors ($n_x$, $n_y$, $n_z$) are provided. In addition to these 6 geometric parameters,
the three flow conditions parameters ($M_\infty$, $AoA$, $p_i$)  are given.
 Combining these 9 numbers, yields an input tensor {\bf X} of dimensions [$n_p \times n_f$, 9].
The corresponding output is described by a tensor {\bf Y} of dimensions [$n_p \times n_f$, 4], where the four columns respectively include
the $Cp$, $Cf_x$, $Cf_y$ and $Cf_z$ values. The described {\bf X} vector provides the relevant inputs for pointwise
regressors that predict the local outputs from local geometrical inputs and flow conditions.
For  regressors predicting complete wall
distributions, relevant {\bf X}$_g$ of size [$n_f$, 3]
may be easily extracted from $X$  (the geometric data being not needed in this case).
Similarly, the corresponding wall distributions {\bf Y}$_g$ of size [$n_f$, $n_p$,4]
 are obtained by a simple reordering of {\bf Y}.
 \\
 
\begin{figure}
\centering
\includegraphics[width=.95\linewidth]{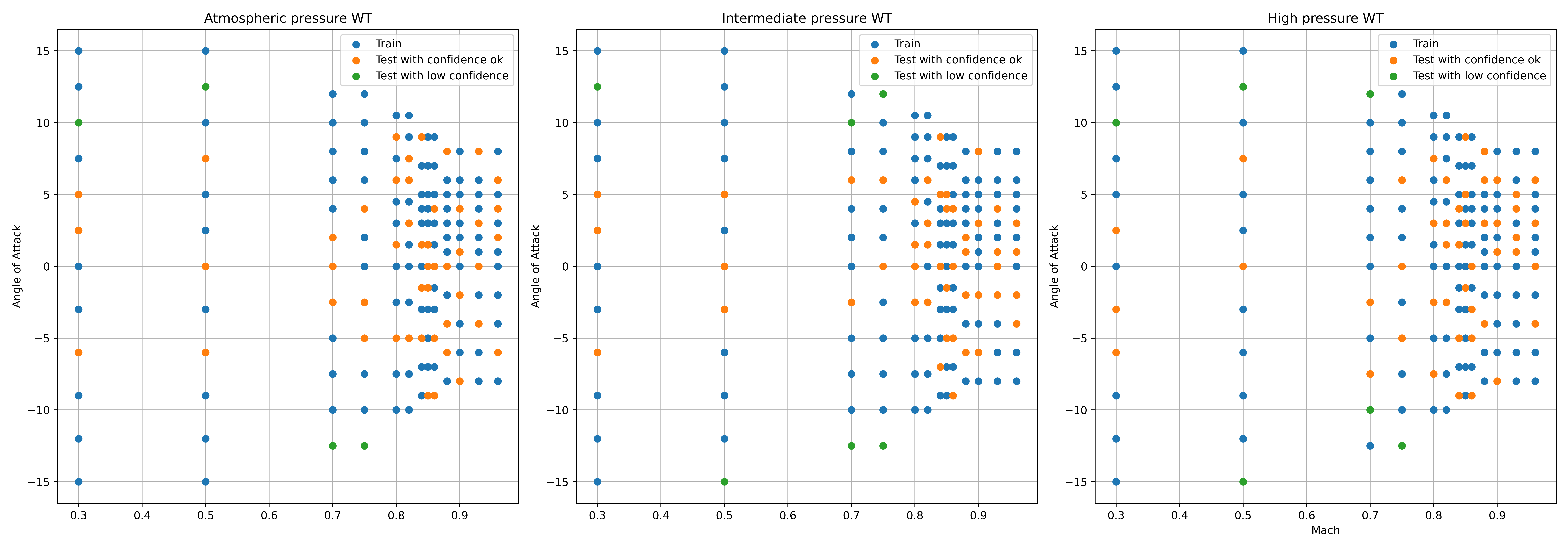}
\caption{Visualization of the train and test repartition among the 468 different computations of the database}
\label{fig:train_test}
\end{figure}

To perform a classical machine learning exercise, the data have been split into a train and a test set.  
The data of one aerodynamic conditions are all together put either in the training or in the testing test for consistency
 with global methods and with practical applications. 
The chosen partition let  $n_{tr}=2/3\times n_f$ conditions (i.e. 312) in the train set and
 $n_{te}=1/3\times n_f$ conditions (i.e. 156) in the test set -- see Figure \ref{fig:train_test}. The split
is done quasi-randomly: for each ($M_\infty$, $p_i$), 4 angles of attack have been choosen
randomly among the 12 to be in the
test set, and the remaining 8 are part of the training set. A minor exception
is made for  $M_\infty= 0.3,$ $M_\infty=0.82$ and $M_\infty= 0.96$
(for all $p_i$ values) where the two extreme angles of attack are forcibly included
in the train set to limit extrapolation by the regressors.
\\
 \indent  For methods involving numerous hyperparameters like neural networks,
  the training set is classicaly split in a inner-training set 
  and a validation set used to optimize the hyperparameters and avoid overfitting.
  For the work reported in \S4 and \S5, a 75\% -- 25\%  partition has been fixed.
  This fraction is in  the range of those of classical studies and courses \cite{HinBek_23,GioBorBen_11,mnist,GooBenCou_16}
  but another choice can be done by each user of the database.

To judge the quality of the prediction provided by a model, two metrics have been chosen.
The first is a $R^2$ score that has been slightly modified to account for the reliability of the simulations. It is meant to assess
the global accuracy of the regressors whereas the second metric aims to characterize the worst possible prediction.
These criteria apply naturally to the outputs of the global regressors predicting complete snapshots $\bf{y}$
 of one output variable.
They are presented hereafter in this context and their adaption to pointwise regressors
 is described in the beginning of next section. The adapted $R^2$ score reads 
\bea
  R^2_y= 1 - \frac{\sum_{i=0}^{n_p\times n_{te}}~~ w_f (y_{i}-\hat{y_{i}})^2}{\sum_{i=0}^{n_p \times n_{te}} ~~ w_f (y_{i}-\overline{y})^2}
  \qquad y \in \{Cp, Cf_x, Cf_y, Cf_z\}
  \label{crit1}
\eea
where a weight $w_f$ is associated to each test flow and $\overline{y}$
the mean of the variable over all locations and flow conditions.
This weight is equal to 1 for all flow conditions with angles of attack
$AoA \in]-10^o,10^o[$.    It is equal to 0.5 if $AoA \leq-10$ or $AoA \geq 10$ to lower the influence of the RANS \& SA
    flows which convergence is expected to be less satisfactory. 
 \\
\indent The second criterion measures the worst performance of the model on the test
conditions: on each flight condition, the mean absolute error is computed and normalized by the mean of the absolute
value of the variable.
Then the maximum of this relative mean absolute error is taken on all the test conditions but those to which a weight
 $w_f$ strictly lower than one has been associated. The size of this reduced test set  is $n_{rte}=140$.
\beas
  rMAE^f_y =\frac{ \sum_{i=0}^{n_p} |y_i^f-\hat{y_i}^f|}{ \sum_{i=0}^{n_p} |y_i^f|} 
  \qquad \qquad f\in [1,n_{rte}]~~~y \in \{Cp, Cf_x, Cf_y, Cf_z\}
\eeas
\bea
  wrMEA_y = \max_{f\in [1,n_{rte}]} rMAE^f_y \qquad   \qquad y \in \{Cp, Cf_x, Cf_y, Cf_z\} 
  \label{crit2}
\eea
The $R^2$ (resp. $wrMEA$) of a regressor is formed finally by averaging the $R^2_v$ (resp. $wrMEA_v$) of the
four output variables.
The regression challenge organized by ONERA will be hosted on the Codabench platform \cite{codabench} --
 https://www.codabench.org/ -- and opened to the public under the name ONERA 468 CRM. 
It will requires participants to submit the wall distributions of the $(Cp, Cf_x, Cf_y, Cf_z)$ at
the conditions of the test set. The solutions will be
evaluated using the two indicators presented above $R^2$ and $wrMAE$.
%
%
%
\section{Performance of classical pointwise regressors}
Pointwise regressors can compute several ouptuts of interest, locally, from pointwise geometrical data (here the 3D coordinates and the local
normal vector of the skin point) and the flow conditions. The regressors are trained minimizing a loss involving all outputs like  
$$
     \sum_{y\in\{Cp, Cf_x, Cf_y, Cf_z\}}    \sum_{i=0}^{  n_p\times n_{tr}}  (y_{i}-\hat{y_{i}})^2
       $$
which makes sense after the output variables have been scaled. After the training, the  $R^2_y$ score and  the $wrMEA_y$ score are 
 computed by formulas (\ref{crit1}) , (\ref{crit2}) before there mean over the four variables, $R^2$ and $wrMEA$, are calculated.
 %
 %
\subsection{Pointwise MultiLayer Perceptron}
A classical Multi-Layer Perceptron (MLP) is known, by its stacking of multiple hidden layers, to be able
to approximate a wide range of functions, making it suitable for nonlinear
regression tasks such as predicting forces on aircraft skin, as explored in this
study. Among the many hyperparameters of a MLP, the ADAM optimizer, the relative batchsize, the learning rate
 (as function of the number of epochs) 
and two regularization techniques (dropout and L2 weight decay that help prevent overfitting) are
selected from previous studies.
The remaining hyperparameters—number of layers, neurons per layer, and activation function—were determined
through Bayesian optimization using the Tree-structured Parzen estimator algorithm implemented in the Optuna framework
\cite{optuna_19} \cite{Wat_23}, with the $R^2$ score on the validation set as the objective function.
This process led to the selection of the LeakyReLU activation function and a 5-hidden-layer network with
(166, 235, 248, 81, 72) neurons in each layer, respectively.
The $R_v^2$ and $wrMAE_v$ scores are presented in Tables \ref{tb06} and  \ref{tb07}  with values of 0.935 and 0.277 respectively.
Additionally, the strategy of tuning a separate MLP network for each output variable was tested but did not
provide substantial enhancements.
%
 %
\subsection{$\lambda$-DNN neural network }
\begin{figure}
\centering
\includegraphics[width=.95\linewidth]{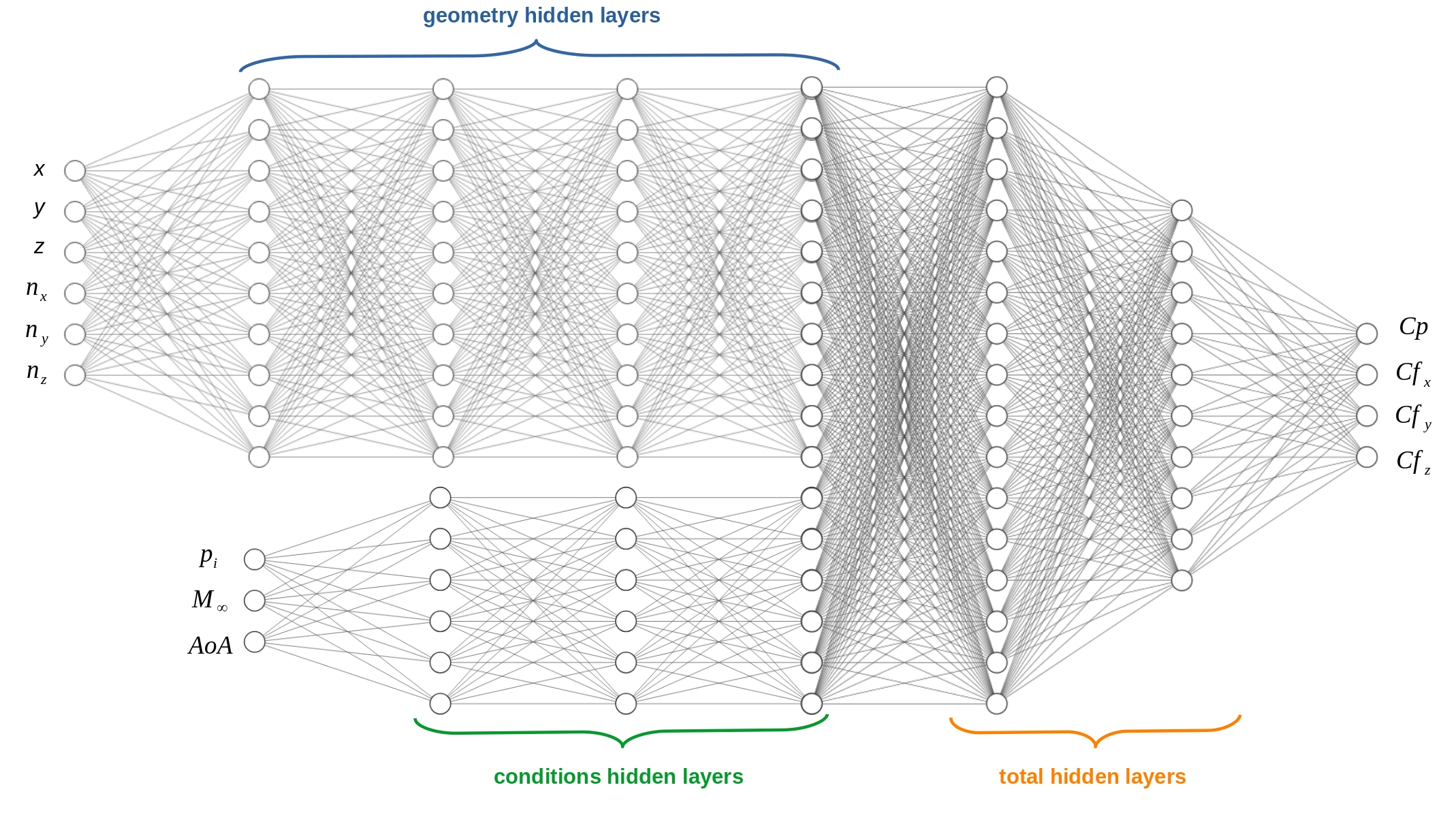}
\caption{Strucutre of a notional $\lambda$-DNN}
\label{fig:lambda-DNN}
\end{figure}

In this section, we present the implementation and evaluation of a specialized MLP architecture, the $\lambda$-DNN,
for the pointwise regression task, following the approach in \cite{KonKapBak_22}. A key feature of this MLP is its
dual-branch input structure, processing each variable type independently before their integration in the network final
fully connected layers. This separation into two branches per type of variable could potentially allow the model to learn
more specialized feature representations for each data category before their combined processing. Figure
\ref{fig:lambda-DNN} provides a visual representation of this architecture. For our application, we separate
the input into two branches: geometric parameters $(x,y,z,n_x,n_y,n_z)$ and aerodynamic conditions
$(M_{\infty}, AoA, p_i)$. The same hyperparameter optimization strategy was employed using the Optuna framework
\cite{optuna_19}. It resulted in a $\lambda$-DNN with the following hidden layer configurations: (107, 116, 236, 139) for
the geometric branch, (240, 179, 114) for the aerodynamic conditions branch, and (230, 162, 124) for the total
(merged) branch.
The $R_v^2$ and $wrMAE_v$ scores are presented in Tables \ref{tb06} and \ref{tb07}, with values of 0.935 and 0.3 respectively. They show little difference
from the scores of the simpler MLP presented above. In fact, the $\lambda$-DNN network can be considered a special MLP where some weights are
forced to zero, so that some layers are not fully connected to each other.

%
%
\subsection{Decision Tree}
Classification and regression trees were initially developed by Breiman et al. \cite{BreFriOls_84}.
A regression tree partitions the data
by iteratively performing tests on the input features. The predicted value in a  terminal node is defined as the mean of
those of the training set
guided to this node by the sequence of tests. The algorithm that has been run \cite{scikitlearn} defines the relevant
successive partitioning
tests by minimizing the squared error between the known outputs and the values in the two created leaves (that is
the mean of the outputs
of the data respectively oriented to these leaves). As in previous studies, whatever the subset of the available
data, the best prediction is obtained when
authorizing very deep trees and terminal leaves with only one sample. For the problem at hand, this creates a tree of depth 108.
Requesting minimum number of some tenths data in terminal leaves
only marginally reduces the accuracy
of the regression which we understand as an indication of regularity of the data.
\\
\indent Considering the  influence of the above menionned parameters is univoque, a final regression tree has
been build with all training and validation
data and no limitation in complexity (no upper bound in depth, no lower bound in samples in leaf).
The resulting $R^2$ value is 0.938 but the worst relative error are quite high
ranging from 0.298 to 0.405 depending on the output -- see Table \ref{tb07}.  
%
%
\section{Performance of classical global regressors}
\subsection{Modewise MultiLayer Perceptron } 
\begin{figure}
\centering
\includegraphics[width=.99\linewidth]{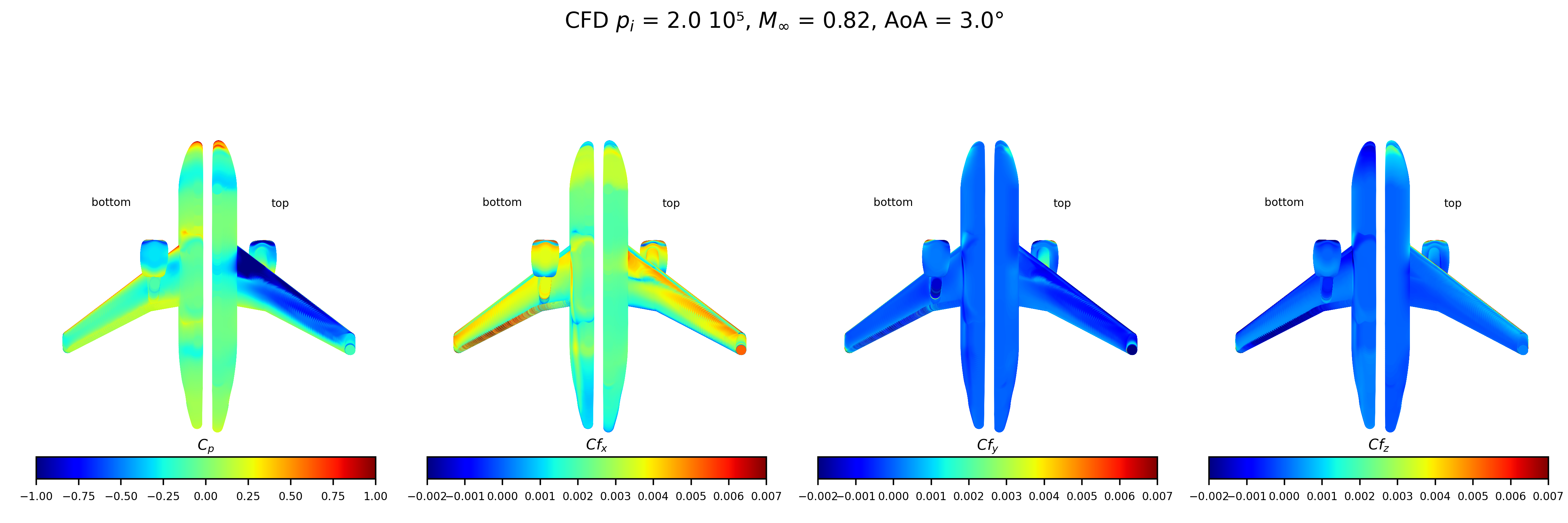}
\includegraphics[width=.99\linewidth]{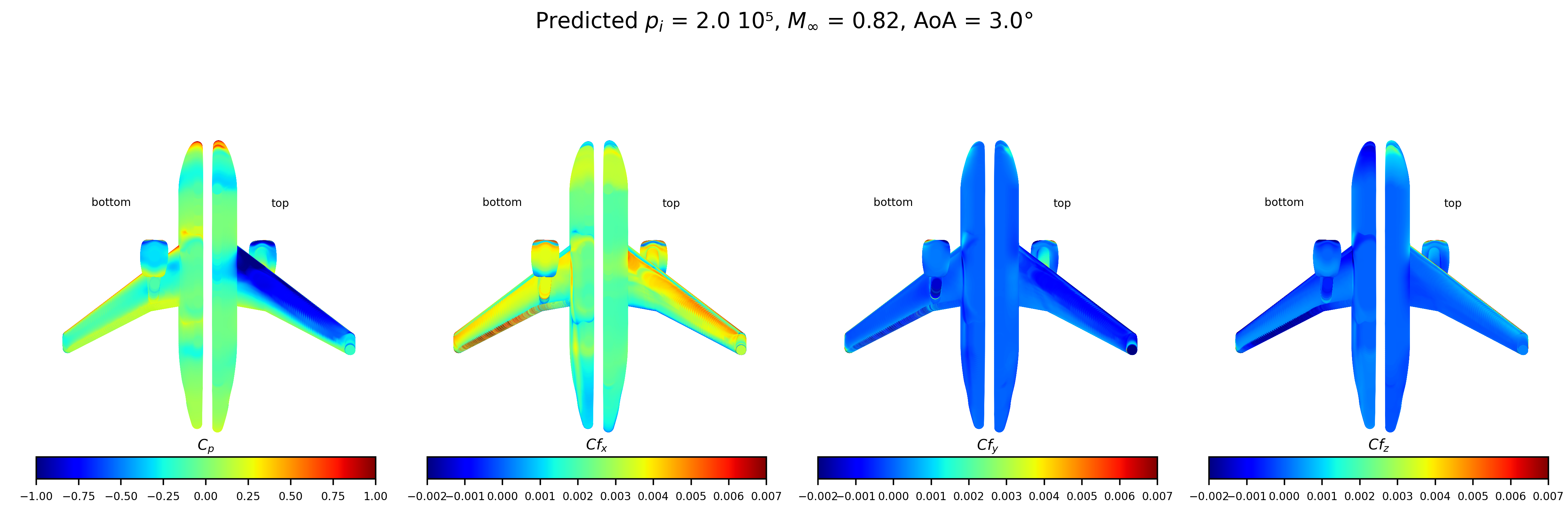}
\caption{Comparison between the CFD fields (top) and the ones predicted by the MLP global network (bottom)}
\label{fig:compareFields}
\end{figure}

Although probably less classical than than its pointwise version, (MLP) can also be defined to predict a complete parietal
field. Provided the field size is about $10^5$, the number of parameters is very important due to the connections between
the penultimate layer (ie the last hidden layer) and the output layer but the calculation is sustainable with current
CPU nodes. A field-by-field approach has been selected to contain the issue of parameter numbers. The classical
traing/validation approach led to a network with five hidden layers of size (75, 120, 1226, 16490).
The $R_v^2$ and $wrMAE_v$ scores are presented in Tables \ref{tb06} and \ref{tb07}. The comparison between
the fileds obtained by CFD and the ones predicted by this model are shown in Figure \ref{fig:compareFields}
for the conditions $M_{\infty} = 0.82$, $AoA = 3.0°$ and $p_i = 2~\times 10^5$.
This model obtains the best score among the seven tested models, both for the overall accuracy $R_v^2$ and for the worst relative error.
A possible explanation could be that it is by far the model with the largest number of parameters, allowing a better generalization.
Despite these relatively good scores, it can be seen in Figure 9 that some detailed characteristics of the flow are still poorly predicted by the model.
For example, the shock appearing on the upper surface of the wing at the considered conditions does not appear in the model prediction ($Cp$ and $Cf_x$)

%
%
\subsection{Neareast Neighbors (kNN) input space interpolation}
This basic method consists in computing a wall distribution $\mathbf{y^*}$ for an unseen flow conditions $\mathbf{p}^*$,
by performing a linear combination of the $\mathbf{y}^f$ 
 snapshots corresponding to the $k$ nearest neighbors of $\mathbf{p}^*$ in the
 flow conditions space:
\begin{equation}
    \mathbf{y^*} = k_{NN}(\mathbf{p}^*) = \frac{\sum_{f \in \mathcal{N}(\mathbf{p}^*)} \varphi_f \mathbf{y}^f}{\sum_{f \in \mathcal{N}(\mathbf{p}^*)} \varphi_f}.
    \label{eq:knn}
\end{equation}
In equation (\ref{eq:knn}), $\mathcal{N}(\mathbf{p}^*)$ denotes the indexes of
the $k$ nearest neighbors of $\mathbf{p}^*$, and $\varphi_f$  the inverse of 
$||\mathbf{p}^f- \mathbf{p}^*||$, the Euclidean distance between $\mathbf{p}^* $ and one of the 
nearest points, $\mathbf{p}^f$, of the training set. (The flow conditions have been first scaled 
to avoid any distortion between the input variables in $||\mathbf{p}^f- \mathbf{p}^*||$.)
 This method does not  account for any non linearity
  of $\mathbf{y}$ as a function of $\mathbf{p}$ and is only accurate (and even exact) 
  if $\mathbf{y}$ is locally a linear function of $\mathbf{p}$, which is definitely not
  the case when the discrete RANS equations are involved to derive $\mathbf{y}$ from $\mathbf{p}$.
  \\
\indent  The accuracy of the method  is quickly decreased when less snapshops are available for the linear combination
  (\ref{eq:knn}) and a large validation set is hence not well-suited in the specific case.
  For this reason, the best number of neighbors
  $k$ is searched by repeatedly removing only ten 
   randomly selected wall distributions, and measuring the accuracy of the prediction for $k\in\{2,3,...12\}$.
  The best number of neighbors is fund to be $ k = 7$ for $Cp$,  $k = 6$ for $Cf_x$, $k$ = 9 for $Cf_y$ and $k = 6$ for $Cf_z$
  but very close results are obtained with these values plus one or minus one. The $R^2_y$ values
  range from 0.921 for $Cf_x$ to 0.954 for $Cp$ but the worst prediction among the wall distributions are very bad
    with $wrMAE$ larger than 0.4 for $Cf_y$ and $Cf_z$. 
%
%
\subsection{Proper Orthogonal Decomposition plus RBF Interpolation}
Proper Orthogonal Decomposition (POD) \cite{Sir_87} is tested here as it is a standard dimensionality reduction
technique in fluid mechanics and although it is not expected to perform excellently when the snapshops
 involve sharp structures. The training wall distributions minus their mean are denoted $\mathbf{\overline{y}}^f ~~
  f\in\{1...n_{tr}\}$. 
 A  Singular Value Decomposition (SVD) of their matrices is first performed:
\begin{equation}
    [\mathbf{\overline{y}}^1,\mathbf{\overline{y}}^2 ...~\mathbf{\overline{y}}^{n_{tr}}] = \mathbf{U}  \Sigma \mathbf{V}^T,   
\label{eq_pod1}
\end{equation}
where $\Sigma$ is a $(n_p,n_{tr})$ matrix with only diagonal and positive non-zero entries, and $\mathbf{U}, \mathbf{V}^T$
are two orthonormal matrices of respective sizes $(n_p,n_p)$ and $(n_{tr},n_{tr})$.
Subsequently, $\Sigma$ is approximated by $\Sigma_r$ keeping only its $r$ largest eigenvalues where $r$ is chosen
 such that the sum of the remaining eigenvalues exceeds
 a threshold (0.99 in this study) times the corresponding complete sum. 
 Provided the eigenvalues of $\Sigma$ were sorted in decreasing order in (\ref{eq_pod1}),
  only $r$ first columns of  $\mathbf{U}$ and $\mathbf{V}$  
  are involved in the resulting approximation of the training data  
\begin{equation}
  [\mathbf{\overline{y}}^1,\mathbf{\overline{y}}^2 ...~\mathbf{\overline{y}}^{n_{tr}}]
  \simeq  \mathbf{U}_r \Sigma_r \mathbf{V}_r^T = \sum_{k=1}^r \sigma_k \mathbf{u}_k \mathbf{v}_k^t  .
\label{eq_pod2}
\end{equation}
Regarding $\mathbf{U}_r$ as a reduced basis for the expression of the wall fields, the $i-th$ coordinates of
the $n_{tr}$ training snapshops  are equal the  to $\sigma_i$ times the elements of
the $i$-th line of $\mathbf{V}_r^T$
($i$-th column of $\mathbf{V}_r$).
This $r$ coordinates are learnt by $r$ regressors as functions of the flow parameters $\mathbf{p}$ of the training set.
The vector of their predictions $\mathbf{z}^*$ for an unseen flow condition $\mathbf{p}^*$, yields the wall-field prediction 
 \begin{equation}
   \mathbf{\overline{y}}^* = \mathbf{U}_r \Sigma_r \mathbf{z}^* . 
      \label{eq_pod3}
 \end{equation}
 
 The SVD from scikit-learn framework \cite{scikitlearn} is completed here by Radial Basis
 Function (RBF) regressors of a well-established numerical toolbox \cite{SMT}.
 The 99\% criterion for the definition of  $\Sigma_r$  yields rather large vector bases
 with respect to the complete training set,
 namely $r = 261 $ for $Cp$,$r = 277 $ for $Cf_x$, $r = 268 $ for $Cf_y$,$r = 256 $ for $Cf_z$.
 The global accuracy of the method is the worst among the considered regressors. 
%
%
\subsection{IsoMap plus RBF interpolation}
IsoMap \cite{TenSilLan_00}
analyses the set of ${\mathbf y}$ distributions as points of a manifold $\mathcal{Y}$ in $\mathbb{R}^{n_p}$.
In this approach, the relevant distances between the ${\mathbf y}$ distributions are hence the geodesic distances on the
manifold that are numerically estimed in two steps: First a nearest neighbors graph of the data ``points''
$\{\mathbf{y}_i \}$ is build;
Then, the  geodesic distances $D_{G_{ij}} = d_G(\mathbf{y}_i, \mathbf{y}_j)$ between data on $\mathcal{Y}$ are
approximated as the shortest path in the graph, using methods such as the Dijkstra algorithm. 
 In a third step, classical Multi Dimensional Scaling (MDS) \cite{Tor_52} is applied to the matrix $\mathbf{D}_G$:
A set of $n_{tr}$ vectors  {\bf z} in $\mathbb{R}^{n_{tr}}$ is defined from the requirements that (a) each components of their
mean is zero (b) the Euclidean distances between them are the same as those stored in $\bf{D_{G}}$. 
The Gram matrix $\mathbf{B}$ of the {\bf z} vectors
is uniquely derived from these two properties
$$    \mathbf{B} = -\frac{1}{2}\mathbf{H} (\mathbf{D}_G \odot \mathbf{D}_G) \mathbf{H} \qquad  \mathbf{H} = \mathbf{I}_{n_{tr}} - {n_{tr}}^{-1} \mathbf{J}$$
$\mathbf{J}$ being a $(n_{tr},n_{tr})$ matrix of ones. The identification of the $n_{tr}$ vectors $\mathbf{z}$
in  $\mathbb{R}^{n_{tr}}$
results from the diagonalization of the symmetric matrix $\mathbf{B}$, $\mathbf{B}=P \Lambda P^T$
and the $r$ dimensional embedding constists in selecting
the $r$ first lines of $ P \Gamma^{1/2}$ instead of all lines. MDS guarantees that the resulting distance matrix
$\mathbf{D}_Z$,
whose entries are given by $D_{Z_{ij}}=||\mathbf{z}_i- \mathbf{z}_j||$, is the best rank $r$ approximation of
$\mathbf{D}_G$ (meaning,  it minimizes the Frobenius norm
 $\rVert \mathbf{D}_G - \mathbf{D}_z \rVert_F$). This is the origin of the name "Isomap":
the algorithm learns a discrete low dimensional embedding $\{\mathbf{z}\}$ of the $\{\mathbf{y}\}$ on $\mathcal{Y}$
that is as isometric as possible with respect to the original manifold.
\\
\indent As in previous section, the coordinates of the embedded variables $\mathbf{z}$, as functions of the flow parameters
$\mathbf{p}$, are learnt using Radial Basis Function \cite{SMT}. The IsoMap algorithm does not go with a backmapping.
In this study kNN interpolation is used to predict the output wall-distribution from the latent space.
\\
\indent First, the size of the latent space is defined by observing the decrease in
$\rVert \mathbf{D}_G - \mathbf{D}_z \rVert_F$ when increasing $r$. As expected, a strong decrease is observed for
$Cp$ as well as for the $Cf$ components going from $r=2$ to $r=3$, the actual dimension of the manifold
$\mathcal{Y}$ in $\mathbb{R}^{n_p}$ (the three aerodynamic variables being independant). After this parameter has been fixed,
the training/validation approach is used again to define the number of neighbors in the kNN approximation.  
We obtain $k = 7$ for $Cp$,  $k = 7$ for $Cf_x$, $k$ = 12 for $Cf_y$ and $k = 9$ for $Cf_z$. The final regression for
 the test set provides results among the most accurate.
%
%

\begin{table}[htbp]
\begin{center}
\begin{tabular}{|l||c||c|c|c|c|c|} 
 \hline
 \hline
               &~~~ $R^2$~~~  & $R^2_{cp}$ & $R^2_{Cfx}$ &   $R^2_{Cfy}$ &  $R^2_{Cfz}$  \\ 
 \hline
 \hline
               \multicolumn{6}{|c|}{Pointwise regressors} \\
\hline
 \hline
  MLP                & 0.935 & 0.974 & 0.927  & 0.931  &  0.910    \\
\hline
  $\lambda$-MLP      &  0.935  &  0.972  & 0.926  &  0.929    &   0.910  \\ 
\hline
  Decision Tree      &  0.938  &  0.960  & 0.924  &  0.934    &   0.934   \\
\hline
\hline
                 \multicolumn{6}{|c|}{Global regressors} \\
 \hline
 \hline
  MLP           & 0.956  &  0.972 &  0.944   &  0.951   & 0.957    \\
\hline
  input-space kNN   & 0.932  &   0.954 & 0.921  &   0.926   &      0.926      \\
\hline
  POD+RBF         &  0.890  & 0.919  & 0.870    & 0.878     &   0.894  \\
\hline
  IsoMap+RBF      & 0.940  & 0.973  & 0.926    &  0.926    &  0.935   \\
\hline
\hline
\end{tabular}
\end{center}
\caption{$R^2$ global regression scores}
\label{tb06}
\end{table}

\begin{table}[htbp]
\begin{center}
\begin{tabular}{|l||c||c|c|c|c|c|} 
 \hline
 \hline
               & $wrMAE$ & $wrMAE_{cp}$ & $wrMAE_{Cfx}$ &   $wrMAE_{Cfy}$ &  $wrMAE_{Cfz}$  \\
 \hline
 \hline
                 \multicolumn{6}{|c|}{Pointwise regressors} \\
 \hline
\multicolumn{1}{|l||}{ MLP }                & 0.277   & 0.205  & 0.197    & 0.386     &  0.318   \\
\multicolumn{1}{|l||}{ \scriptsize{$(M_\infty,AoA,10^{-5}p_i)$} }                       &    & (0.3, -6$^o$, 1)  &  (0.85, 9$^o$, 4)    &  (0.85, 9$^o$, 4)     &  (0.88, 8$^o$, 4)    \\
\hline
 \multicolumn{1}{|l||}{$\lambda$-MLP }     & 0.300   &  0.233 & 0.207    &  0.399    & 0.362    \\ 
 \multicolumn{1}{|l||}{\scriptsize{$(M_\infty,AoA,10^{-5}p_i)$}  }                      &    & (0.3, -6$^o$, 1)  & (0.85, 9$^o$, 4)    & (0.90, -8$^o$, 4)      &   (0.90, -8$^o$, 4)   \\ 
\hline
\hline
  \multicolumn{1}{|l||}{ Decision Tree }     & 0.325   &  0.307 &  0.289    &  0.405    & 0.298    \\
 \multicolumn{1}{|l||}{ \scriptsize{$(M_\infty,AoA,10^{-5}p_i)$}   }                    &   &  (0.3, -6$^o$, 1) &  (0.85, 9$^o$, 4)   & (0.88, 8$^o$, 4)     & (0.3, -6$^o$, 1)    \\ 
\hline
\hline
                 \multicolumn{6}{|c|}{Global regressors} \\
 \hline
 \hline
\multicolumn{1}{|l||}{  MLP }                   & 0.228   & 0.198  & 0.163    & 0.314     &  0.237   \\
\multicolumn{1}{|l||}{ \scriptsize{$(M_\infty,AoA,10^{-5}p_i)$} }                       &    & (0.3, -6$^o$, 1)   & (0.85, 9$^o$, 4)    &  (0.3, 2.5$^o$, 1)    &  (0.90, -8$^o$, 4)   \\
\hline
\multicolumn{1}{|l||}{  parameter-space kNN }   &  0.346  &  0.332  &  0.194  &  0.445   &  0.411  \\
\multicolumn{1}{|l||}{ \scriptsize{$(M_\infty,AoA,10^{-5}p_i)$} } &    &   (0.3 ,-6$^o$, 4) &  (0.96,-2$^o$, 2)   &  (0.3, -3.$^o$, 4)    &   (0.5, 0.0$^o$, 2)  \\
\hline
 \multicolumn{1}{|l||}{ POD+RBF }                &  0.292  & 0.291  &  0.166   &   0.399   &  0.310   \\
\multicolumn{1}{|l||}{  \scriptsize{$(M_\infty,AoA,10^{-5}p_i)$} }      &    &  (0.3 ,-6$^o$, 4)   & (0.85, 9$^o$, 4)  &  (0.3 ,-6$^o$, 4)     &    (0.3 ,-6$^o$, 4)      \\
\hline
 \multicolumn{1}{|l||}{ IsoMap+RBF }           &  0.296  & 0.210   & 0.224    & 0.392     &  0.356   \\
\multicolumn{1}{|l||}{  \scriptsize{$(M_\infty,AoA,10^{-5}p_i)$} }  &    & (0.3, -6$^o$, 4)  &   (0.5, 7.5$^o$, 4)   &  (0.3 -3$^o$,   4) &   (0.3, 2.5$^o$, 1)   \\
\hline
\hline
\end{tabular}
\end{center}
\caption{$wrMAE$ regression scores}
\label{tb07}
\end{table}

%
\section{Conclusion}
A surprisingly large number of regressors of various types is currently used by the community to study series
of wall distributions extracted from CFD calculations and to build regressors to generalize these data
\cite{YonAndVal_18,LiDuMar_22,And_24}. In an attempt to facilitate scientific activities on this topic, ONERA has
computed a series of 468 flows about the wing-body-polyn-nacelle CRM with the accuracy required
by Applied Aerodynamics. A 2/3 (training and validation) 1/3 (test) split of the data supports
a regression challenge that will soon
be proposed to the community. A first series of numerical experiments on this challenge with both, pointwise and
 global regressors, points out its intermediate difficulty:
 it is possible to obtain globally statisfactory accuracy for the test conditions but, conversely,
 the error on the worst predicted flow was always high in our trials. We hope that this challenge will be a useful
  support for future cooperative activities.
  \section*{Funding, Acknowledgments}
  This work was founded by ONERA's internal ressources. \\
  The authors thank Samir Beneddine for his fruitful recommendations and Antoine Jost for his manuscript review.
  

%
\bibliography{biblio-ML}{}
\bibliographystyle{elsarticle-num.bst}

\appendix

\end{document}